\documentclass[runningheads]{llncs}
\usepackage{graphicx}
\usepackage{booktabs}
\usepackage{comment}
\usepackage{adjustbox}

\usepackage{hyperref}
\usepackage[style=nejm, citestyle=numeric]{biblatex}
\usepackage{authblk}

\usepackage{comment}
\graphicspath{{./figs/}}
\usepackage{balance}
\usepackage{xspace} 
\usepackage{enumitem}
\usepackage{multirow}
\usepackage{mathtools}
\usepackage{pifont}

\usepackage{tikz}
\usepackage{wasysym}
\usepackage{textcomp}
\usepackage{algorithmic}
\usepackage[ruled,vlined]{algorithm2e}
\usepackage{subfigure}
\usepackage{amsfonts,amssymb}
\usepackage{amsmath}

\usepackage{amsthm}

\theoremstyle{remark}
\newcommand{\tabincell}[2]{\begin{tabular}{@{}#1@{}}#2\end{tabular}}

\usepackage{array}

\usepackage[font=footnotesize]{caption}
\author{Yixuan Liu\inst{1} \and
Hu Wang\inst{2} \and
Xiaowei Wang\inst{2} \and
Xiaoyue Sun\inst{2} \and
Liuyue Jiang\inst{2} \and
Minhui Xue\inst{2}}
\authorrunning{Y. Liu et al.}
\institute{Quaintquant Pty Ltd, Adelaide SA 5000, Australia \and
The University of Adelaide, Adelaide SA 5000, Australia}
\begin{document}
%
\title{Empirical Sufficiency Featuring Reward Delay Calibration in Reinforcement Learning}
\titlerunning{Empirical Sufficiency Featuring Reward Delay Calibration in Reinforcement Learning}
%

%
%
\maketitle              

\begin{abstract}
Appropriate credit assignment towards delay rewards is a fundamental challenge for reinforcement learning. To tackle this problem, we introduce a delay reward calibration paradigm from a classification perspective. We hypothesize that well-represented vectors of states share similarities if they contain the same or equivalent essential information. To this end, we define an empirical sufficient distribution, where the state vectors within the distribution will lead agents to environmental reward signals in the consequent steps. Therefore, a purify-trained classifier is designed to sketch the distribution and further generate calibrated rewards. We then examine the correctness of sufficient state extraction by tracking the real-time reward obtaining. The experimental results demonstrate that the classifier could generate timely and accurate calibrated rewards to make the model training process more efficient. We also identify and discuss that the sufficient states extracted by our model resonate with the observations of humans.
\end{abstract}

\section{Introduction}

Reinforcement learning (RL) approaches have made incredible breakthroughs in various domains~\cite{DBLP:journals/nature/SilverHMGSDSAPL16,DBLP:journals/nature/MnihKSRVBGRFOPB15,DBLP:journals/corr/abs-1910-07113,DBLP:journals/nature/VinyalsBCMDCCPE19}, where the performance exceeds human. The reinforcement learning theoretically models sequential decision tasks as dynamic programming processes to maximize expected accumulated rewards. Given that environmental rewards generally cannot entirely reflect the contribution of each action in an episode, existing approaches commit to distributing different credits to individual decisions, known as credit assignment~\cite{DBLP:books/lib/SuttonB98}. Bellman equation-based models calculate state values based on the expectation of gathered rewards from later steps, which unreasonable values are assigned at times. This problem becomes even more intractable when reward signals are extremely sparse or severely delayed.

In this paper, we formulate an purify-trained classification mechanism to extract empirical sufficient conditions of acquiring desired environmental signals, which generally indicate positive rewards. We refer to this extraction formulation as an Empirical Sufficient Condition Extractor (ESCE) to fairly assign calibrated rewards to corresponding states in advance of delayed rewards. We first propose to identify empirical sufficient states with a classification mechanism. To train the classifier with partially labeled data, we label the state vectors with instantly-acquired environmental signals. Then, we proposed to train the classifier with two stages, wherein a novel purified training process is conducted. In addition to existing value-based estimation, the ESCE provides concrete and objective predictions. 

We equip Deep Q-Networks (DQN)~\cite{DBLP:journals/corr/MnihKSGAWR13} and Asynchronous Advantage Actor Critic (A3C)~\cite{DBLP:conf/icml/MnihBMGLHSK16} agents with the ESCE and measure the performance on multiple prevalent games, most of which have delayed discrete rewards. We examine the extraction correctness by tracking the reward obtaining, and further record the accuracy/recall changes of ESCE on the fly. The results show the agents guided by the proposed empirical efficiency achieve significant improvements in convergence, especially in the scenarios with delayed rewards. Furthermore, we constructively modify the environment to render the rewards to be even more delayed, termed as \textit{hindsight rewards} settings. The results show that the calibrated rewards are able to lead agents to acquire well-learned target policies even if in the hindsight rewards scenarios. 
In addition to quantitative experiments, we screenshot the identified sufficient states, showing the high similarity of calibrated rewards received with human's perceptions. Our contributions can be summarized as follows:

\begin{itemize}

\item We introduce a model to extract empirical sufficient conditions from classification perspective, where we propose to significantly reduce the uncertainty with purified training scheme.

\item We define empirical sufficient conditions and formulate a calibrated rewards in line with corresponding environmental signals to tackle the reinforcement learning reward delay issues. Such that, the agents receive rewards whenever the empirical sufficient conditions are satisfied.

\item The experimental results show reward-calibrated agents are able to learn significantly better policies in the scenarios where rewards have been severely delayed, compared with agents without equipping the proposed calibrated rewards. Moreover, further discussion of sufficient states extracted by the proposed model are conducted. It shows the identification of sufficient states resonates with the observations of humans.

\end{itemize}

\section{Related Work}
\subsection{Intrinsic Motivation}

Intrinsic rewards~\cite{DBLP:conf/nips/SinghBC04,ryan2000intrinsic} are inspired by intrinsic motivation to either encourage exploration or fulfil certain purposes. The mechanism to obtain intrinsic rewards is usually independent to it of the environmental rewards. Exploration-oriented intrinsic rewards are generally correlated to the novelty or informative acquisition of new arrival states~\cite{DBLP:conf/icml/PathakAED17, DBLP:conf/iclr/BurdaESK19, DBLP:conf/nips/HouthooftCCDSTA16, DBLP:journals/corr/abs-1903-07400}. Since the awarding mechanism of intrinsic rewards does not depend on environments, as an exchange, the policy learned may mismatch the final objects. In addition to exploration, intrinsic rewards can often be found in hierarchical frameworks~\cite{DBLP:conf/nips/KulkarniNST16, DBLP:conf/icml/VezhnevetsOSHJS17, DBLP:conf/iclr/FransH0AS18}. Also, intrinsic rewards are used to assist agents to learn optimal or near-optimal policies in a more direct manner~\cite{DBLP:journals/corr/abs-2007-10835, DBLP:conf/nips/ZhengOS18, DBLP:journals/corr/abs-1912-05500}. Unlike the exploration encouraged intrinsic reward, the proposed ESCE is able to generate accurate intrinsic rewards by identifying key states leading to learn a better policy.

\subsection{Credit Assignment for Delayed Rewards}

Most expectation estimators in reinforcement learning rely on Bellman equation, where the expectation accumulated rewards are passed through states following the basic idea of dynamic programming~\cite{DBLP:conf/nips/LeeCC19, DBLP:conf/nips/Arjona-MedinaGW19, DBLP:conf/icml/NgHR99, DBLP:conf/aaai/MaromR18}. To make the training more efficient, one method is to build an extra model to capture critical states for better model building~\cite{DBLP:journals/jmlr/SuttonMW16, DBLP:conf/nips/KeGBBMPB18, DBLP:journals/corr/abs-1810-06721}. Ideologically, ~\cite{DBLP:conf/nips/IrpanRBHIL19} introduce binary classification into value estimation and positive-unlabeled learning~\cite{DBLP:conf/nips/KiryoNPS17} is adopted to distinguish promising and catastrophic states. Different from existing works, our work evaluates states by discriminating states as a binary classification problem without relying on Bellman equation and the idea of dynamic programming. Specially, by accurately differentiating states between ``sufficient for success'' and ``insufficient for success'', 
a powerful intrinsic reward estimator is built (refer to Sections~\ref{subsection:proximal labeling} and~\ref{subsection:purification training} for more details).

\section{Delayed Reward Calibration}

\subsection{Empirical Sufficient Distribution (ESD)}
\label{subsection:ESD}

The emergence of a particular state always leads to a consequence, which we refer as sufficient conditions. We consider a set of particular environmental signals as target consequences and let the model predict these consequences through stored state vectors to yield the sufficiency to incur these consequences. We hypothesize that closely distributed state vectors share similar information which will further incur indistinguishable results. We therefore leverage this hypothesis to proceed with the classification-based evaluation.

\theoremstyle{definition}
\begin{definition}{\bf Reward Empirical Sufficiency.} 
\label{definition:1}
If there exists a continuous space such that state vectors within it incur a particular environmental consequences, we define these states as the Reward Empirical Sufficiency.
\end{definition}

Given an environment, let $R_\mathrm{positive}$ be the desired environmental signals, a.k.a. positive rewards; conversely, $R_\mathrm{negative}$ represents undesired environmental signals, including negative rewards, agent's deaths and game endings. We further define Empirical Sufficient Distribution (ESD) as: when agents explore the environment with a specific policy, if an agent reaches a state $s_\mathrm{suf}^\pi$ and invariably acquires $R_\mathrm{positive}$, we define a state $s_\mathrm{suf}^\pi$ as the empirical sufficient state to acquire $R_\mathrm{positive}$. If the all states in a distribution are empirical sufficient states, we then consider the distribution as the Empirical Sufficient Distribution of $R_\mathrm{positive}$.

\begin{figure}[t]
\centering
\scalebox{1}{
\centerline{\includegraphics[width=1\textwidth]{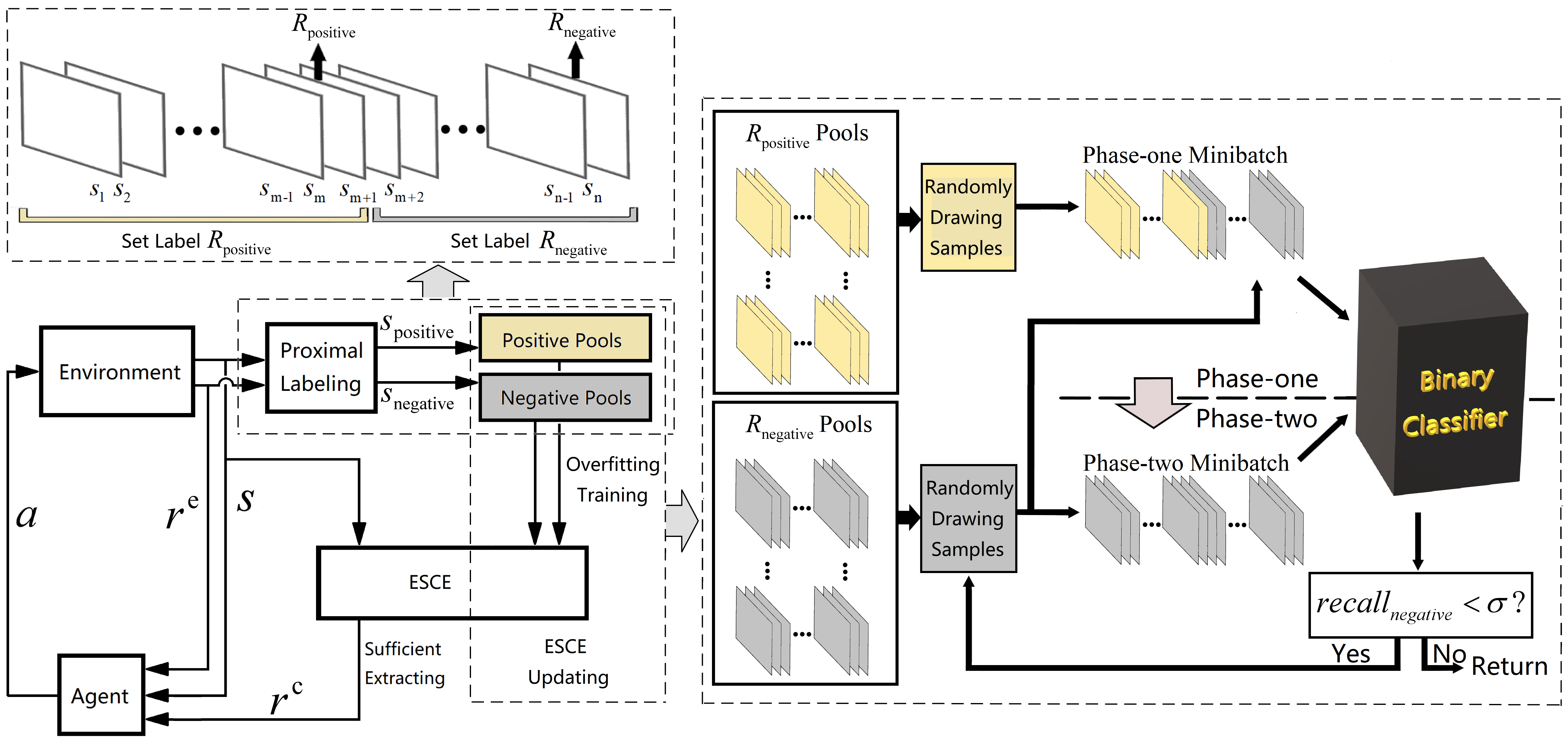}}
}
\caption{The overall framework of a RL agent equipped with ESCE. Agents receive environmental rewards and calibrated rewards from ESCE as the total rewards. 
The ESCE model examines each state and provides calibrated rewards whenever a state is identified as an empirical sufficient state. Within the ESCE training process, state vectors are automatically labeled and then stored in corresponding pools. The ESCE network is updated with purified training, where phase one is a binary classification with data from pools and $R_\mathrm{negative}$ pools, and phase two is the proposed purified training updated with $R_\mathrm{negative}$ data alone.}
\label{fig:architecture}
\end{figure}

\subsection{Learning with Hybrid Reward Functions}

We design Empirical Sufficient Condition Extractor (ESCE) as an independent module that can be incorporated into multiple mainstream reinforcement learning frameworks. Calibrated rewards are provided by ESCE when a state meets the empirical sufficient condition (within the Empirical Sufficient Distribution (ESD)). We denote $\pi(s_t; \theta)$ as the learned policy, where $s_t$ is the observed state at time $t$ and $\theta$ is the set of parameters of the policy network; $r_t^\mathrm{c}$ is the calibrated reward generated by ESCE at time step $t$ and $r_t^\mathrm{e}$ is the environmental reward from the environment at time step $t$. 
The total reward function is synthesized with calibrated signals and environmental signals, $r_t = \alpha r_t^\mathrm{c} + \beta r_t^\mathrm{e}$, where $\alpha$ and $\beta$ are the weight coefficients of corresponding rewards. Our baseline, optimized by environmental rewards, has coefficients $\alpha$ = 0 and $\beta$ = 1. The policy network is optimized to maximize the expected accumulated rewards: 

\begin{equation}
\label{eq:policy}
\pi^*(s; \theta) = \max_{\theta}[\mathbb{E}_{\pi} (\sum_{k=0}^\infty \gamma^k r_{t+k+1}) | s_t=s].
\end{equation}

The overview of our framework is illustrated in Figure~\ref{fig:architecture}. Whenever an empirical sufficient state is identified by ESCE, a positive reward is offered. The calibrated reward is available until receiving an environmental signal. Meanwhile, these states are labeled and stored in the corresponding pools for classifier updating in the future. The training of the ESCE and policy network proceeds alternately.

\begin{algorithm}[tbp]
\SetAlgoLined
 Initialize Extractor network and policy $\theta$\;
 Initialize $R_\mathrm{positive}$ pool and $R_\mathrm{negative}$ pool\;
 \Repeat {
 Converged\
}{
Initialize temporary storage\;
 \While{any state pool is not full}{
  \If{state $S_t$ is identified as $R_\mathrm{positive}$ and no calibrated reward have been given to the agent in this round}{
   Assign calibrated reward to the agent\;
   }{  }
  Push states to a temporary storage\;
  Update the policy network with the parameters $\theta$\;
  \If{environmental signal is not null}{
   Push temporary storage to pool based on positive/negative environmental signals\;
   Clear temporary storage\;
   Reset calibrated awarding status (Start a new round)\;
   }{  }
 }
 Update Extractor with states from both pools\;
 \While{Recall of $R_{\text{negative}} $ \textless$\sigma$}{
  Update ESCE parameters with data from $R_\mathrm{negative}$ pool\;
 }
 Clear all pools\;
}
 \caption{Empirical Sufficient Conditions Extractor (ESCE)}
\end{algorithm}

\subsection{The Labeling of ESCE}
\label{subsection:proximal labeling}

We implement the ESCE training with binary classification and take individual observation of states as inputs. For the labeling process, we self-supervised this process to treat newly-acquired environmental signals as classification labels.

First, we cut a long sequence of episode into several rounds along with instantly-acquired environmental signals. Later on, these environmental signals are adopted as the labels for the states within a round. Specifically, a newly received environmental signal is treated as the label for states starting from the lastly-received environmental signal until this one. The labeling procedure is shown in the Figure~\ref{fig:architecture}. We generalize desired environmental signals as $R_\mathrm{positive}$, and all undesired environmental signals as $R_\mathrm{negative}$. Accordingly, $R_\mathrm{positive}$ and $R_\mathrm{negative}$ pools are created for labeled images collection.

As we adopt nearby environmental signals as labels, a potential problem to be solved is label ambiguity. For some states whose consequences are not determined, their labels may be different in different episodes, each beginning state for instance. As a result, the state vectors of two different labels might be densely mixed, and there might not be a clear boundary between them, which has a significant difference towards normal supervised-learning scenarios. 
In order to resolve the label ambiguity issue, regarding Definition~\ref{definition:1}, ESD only contains states that lead to one particular environmental signal. It indicates that ESD should only include states with unique labels. Therefore, we formulate ESCE training as an purified training process to exclude state space with ambiguous labels.

\subsection{Purified Training Process}
\label{subsection:purification training}

We formulate a two-phase training process to extract purified empirical sufficient distribution. In phase one, we expect ESCE to correctly identify all samples as much as possible by assigning dominant labels to the states since ambiguous states may confuse the classifier. In phase two, to exclude insufficient states, an purified training mechanism is adopted to optimize the decision boundary. We measure performance improvement of ESCE with {\em Precision} and {\em Recall} on $R_\mathrm{positive}$. The calculation of {\em Precision} and {\em Recall} can be presented:

\begin{equation}
\label{eq:precession}
\begin{aligned}
{\displaystyle {\text{Precision}}}_\mathrm{pos}= \frac{N_\mathrm{suff}}{N_\mathrm{ident}},
\end{aligned}
\end{equation}

\begin{equation}
\label{eq:recall}
\begin{aligned}
{\displaystyle {\text{Recall}}}_\mathrm{pos} = \frac{N_\mathrm{suff}}{N_\mathrm{pos}},
\end{aligned}
\end{equation}
where the {\em Recall} indicates the ratio of identification coverage on those samples associated with $R_\mathrm{positive}$, and the {\em Precision} shows how accurate the identification is. We set the number of rounds including identified states as $N_\mathrm{ident}$, and let $N_\mathrm{sufficient}$ denote the number of rounds that contains identified empirical sufficient states leading to $R_\mathrm{positive}$. Additionally, we set the total number of samples leading to $R_\mathrm{positive}$ acquired as $N_\mathrm{pos}$.

\begin{figure*}[tbp] \centering
\scalebox{0.85}{
\subfigure[ ] {
 \label{fig:a}
\includegraphics[width=0.32\columnwidth]{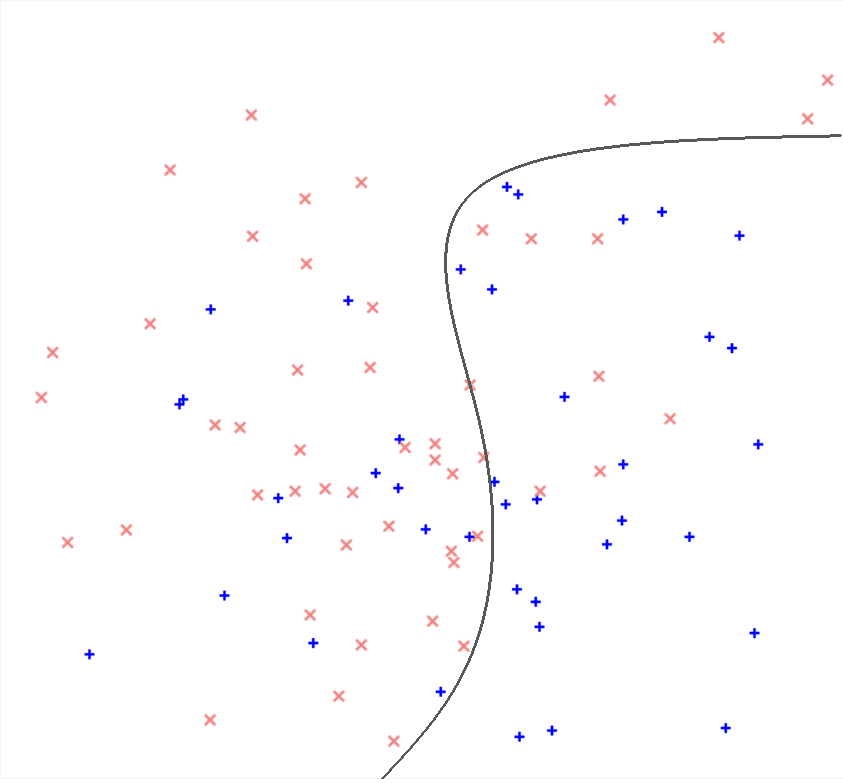}
}     
\subfigure[ ] { 
\label{fig:b}     
\includegraphics[width=0.32\columnwidth]{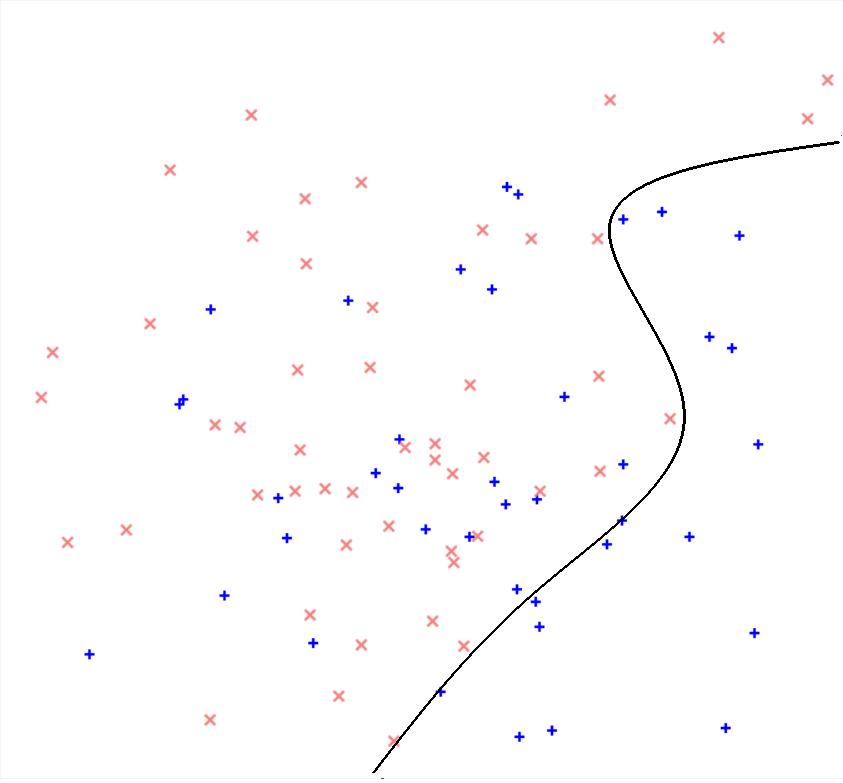}     
}    
\subfigure[ ] { 
\label{fig:c}     
\includegraphics[width=0.32\columnwidth]{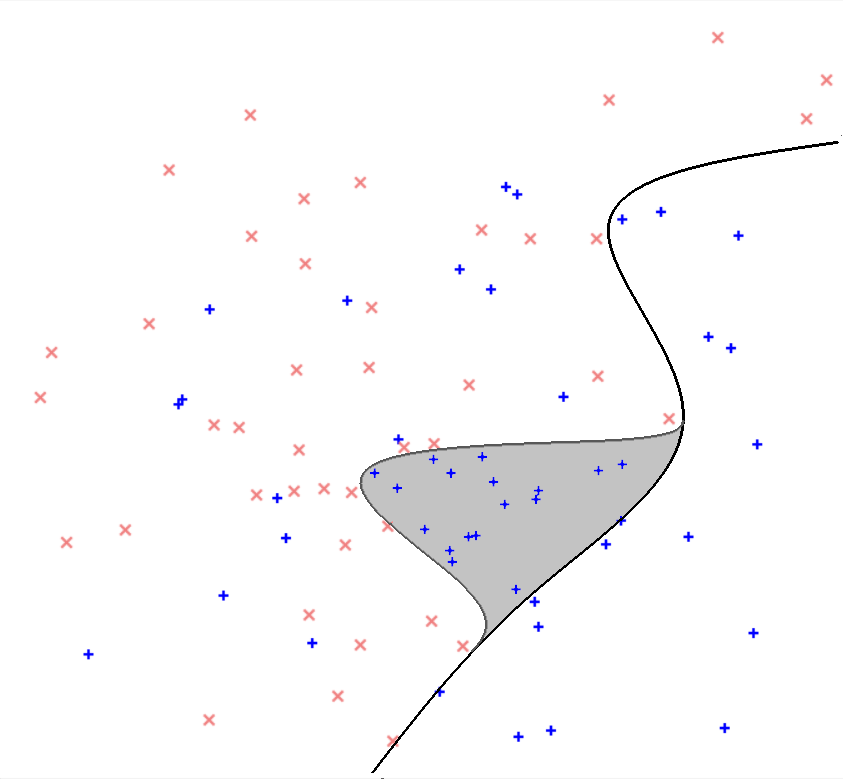}     
}
}
\caption{\ref{fig:a} A diagram of the decision boundary improvement after phase-one of the proposed purified training: the classifier tries to make most of the samples correctly distributed on both sides of the boundary. However, there are still a lot of misclassified samples falsely allocated due to label ambiguity. \ref{fig:b} A diagram of decision boundary after phase-two of the proposed purified training: all state vectors labeled with $R_\mathrm{negative}$ can be correctly classified and it matches the definition of ESD. \ref{fig:c} A diagram of sensitive sampling: the ESD may be changed with the updates of policy. The state vectors inside the grey region were the insufficient states, and turn into empirical sufficient states later on.}
\label{fig}
\end{figure*}

The phase one is carried out by a binary classification training. The training data comes from both $R_\mathrm{positive}$ and $R_\mathrm{negative}$ pools, where all $R_\mathrm{positive}$ samples ($s_\mathrm{pos}$) are labeled with desired signals ($r_\mathrm{pos}$), and $R_\mathrm{negative}$ samples ($s_\mathrm{neg}$) have undesired signal labels ($r_\mathrm{neg}$). Both types of samples are adopted to prudently maximize the {\em Recall} of ESCE (Figure~\ref{fig:a}). Let the function learner $f$ generate estimations of future rewards $\hat{r}$ with parameters $\psi$ of ESCE, where the calibrated rewards $\hat{r}_\mathrm{pos}$ and $\hat{r}_\mathrm{neg}$ are generated from $s_\mathrm{pos}$ and $s_\mathrm{neg}$, respectively:

\begin{equation}
\label{eq:r_pos}
\hat{r}_\mathrm{pos} = f\Big(s_\mathrm{pos};\psi\Big),
\end{equation}
\begin{equation}
\label{eq:r_neg}
\hat{r}_\mathrm{neg} = f\Big(s_\mathrm{neg};\psi\Big).
\end{equation}

Within the optimization to correctly classify each sample, binary cross-entropy objective is adopted. The loss of phase one measures the discrepancy between the estimated rewards and their ground truth, which are defined as follows:

\begin{equation}
\label{eq:PA}
\begin{aligned}
L_1=- r_\mathrm{pos} \cdot \log \left(p\left(\hat{r}_\mathrm{pos}\right)\right)+\left(1-r_\mathrm{neg}\right) \cdot \log \left(1-p\left(\hat{r}_\mathrm{neg}\right)\right).
\end{aligned}
\end{equation}

In phase two, we try to maximize {\em Precision} only with $R_\mathrm{negative}$ samples. The classification boundary is updated to acquire a distribution of purified $R_\mathrm{positive}$ samples by excluding all insufficient samples (refer to Figure~\ref{fig:b}). The objective function adopted in phase-two is defined as follows:

\begin{equation}
\label{eq:IC}
\begin{aligned}
L_2=- \left(1-r_\mathrm{neg}\right) \cdot \log \left(1-p\left(\hat{r}_\mathrm{neg}\right)\right).
\end{aligned}
\end{equation}

After the termination of phase-two, the states recognized as $R_\mathrm{positive}$ should only include states with label $R_\mathrm{positive}$. In other words, all states identified as $r_\mathrm{pos}$ would incur $R_\mathrm{positive}$, in line with the definition of empirical sufficient distributions and thus the Empirical Sufficient Distribution (ESD) is formed. We show the ESCE architecture in Algorithm 1.

\subsection{Sensitive Sampling}

To efficiently update the ESCE network, we adopt a sensitive sampling strategy for purified training. ESD may change with the updates of policy network. For instance, if an policy network is significantly improved, it may stably acquire rewards which were previously unattainable. As a result, the space containing ESD might be expanded. An efficient way to update the parameter $\theta_E$ of ESCE is to pay more attention on those ``hard'' examples (refer to Figure~\ref{fig:c}). Accordingly, we build two extra state pools for data collection. One pool is for miss-identified samples and the other pool is built for false-identified samples. These two state pools force ESCE to focus on those ``hard'' examples for faster convergence and better performance. Empirically, 75\% of training data are imported from two sensitive pools in our experiments.

\section{Experiments Setup}

\subsection{Sparse and Delay Reward Setting}
\label{section: A2}

We first evaluate our model and the comparing models on MountainCar-v0 and then move on to 6 Atari 2600 games of OpenAI Gym ~\cite{brockman2016openai}. Positive rewards are defined as $R_\mathrm{positive}$, while the negative rewards, deaths, and game endings are denoted as $R_\mathrm{negative}$. Reinforcement Learning algorithms are inevitably facing the low sample efficiency issue, while the delayed rewards worsen the problem. Especially, in those games that rewards are offered lately, for instance, in the game Bowling-v0, the environmental rewards are determined when the character throws the ball, the proposed calibrated reward is able to mitigate the problem significantly. The longer gaps in time between empirical sufficient state and reward are, the more challenge the agent has to learn the ''correct'' policies.

\subsection{ESCE on DQN and A3C Architecture}
\label{section: A3}

We examine the performance and universality by attaching to different algorithm frameworks. In MountainCar-v0 experiments, we join ESCE with a DQN agent, where the DQN agents update its policy with calibrated rewards. We compare the performance with another DQN agent awarded by the environment.

For Atari environment, we adopt an A3C-LSTM framework as our backbone architecture. The original RGB image with size 210$\times$160 is converted to 80$\times$80 gray-scale frames. Four continuous frames are stacked as the input. In A3C architecture, four convolution layers and max-pooling layers are adopted. An LSTM layer with 512 units is followed by two heads --- a policy head and a value function head.

\begin{table*}[tp]
    \centering
    \caption{Comparison with baselines on hindsight rewards settings: only the calibrated rewards are provided $r_t^\mathrm{c} = 1\cdot r_t^c + 0\cdot r_t^e$ and rewards are offered after a $R_\mathrm{negative}$ signal or at the end of episodes. The real time {\em Precision} and {\em Recall} of $R_\mathrm{positive}$ are presented in the right side of the figures.}
    \label{table:3}
    \includegraphics[scale=0.3]{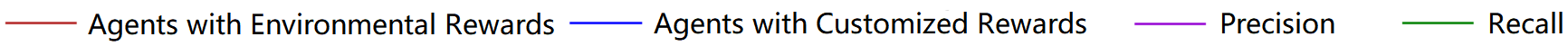}
    \begin{tabular}{m{0.6cm}<{\centering} m{3.6cm}<{\centering} m{3.6cm}<{\centering} m{3.6cm}<{\centering}}
        \toprule
         & \tabincell{c}{\scriptsize FishingDerby-v0} & \tabincell{c}{\scriptsize Breakout-v0} & \tabincell{c}{\scriptsize Pong-v0}  \\
        \midrule
        \tabincell{c}{\scriptsize $r_t^\mathrm{c}$\\ \quad } & 
        \includegraphics[scale=0.14]{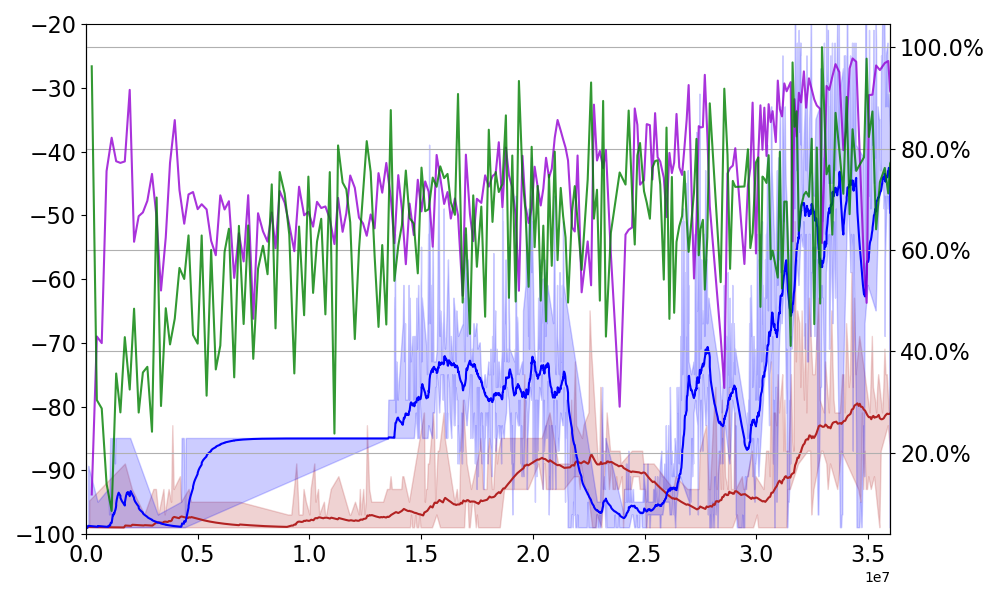} & 
        \includegraphics[scale=0.14]{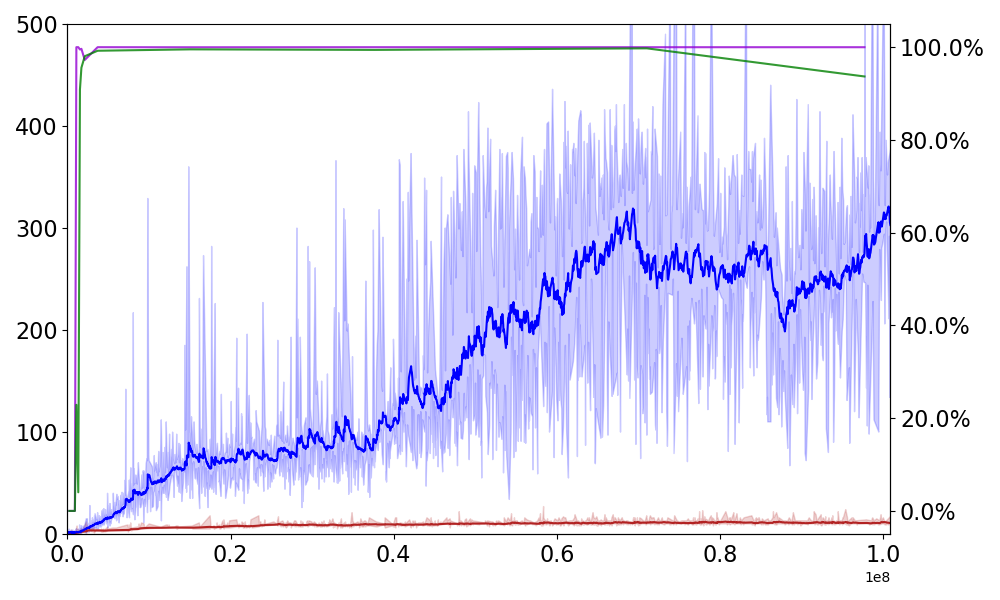} & 
        \includegraphics[scale=0.14]{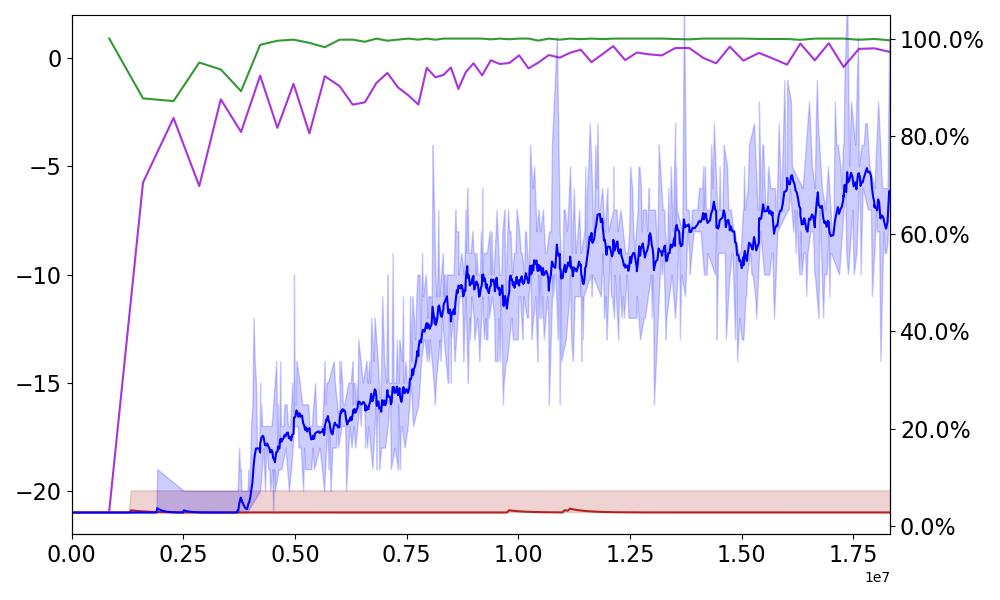} \\
        \midrule
        \tabincell{c}{\scriptsize Max Score } & 
        \tabincell{c}{\scriptsize 5.0  } & 
        \tabincell{c}{\scriptsize 782.0  } & 
        \tabincell{c}{\scriptsize 3.0  } \\
        \bottomrule
    \end{tabular}
\end{table*}

Considering the policy is evolving on the fly, the empirical sufficient conditions should be updated accordingly. To ensure the extracted empirical sufficient distribution is up to date with the policy, we train the ESCE model with data sampled from the latest episodes. Subject to this reason, the maximum capacity of the datasets for the training of empirical sufficient state classifier should be flexibly decided. On the other hand, model-free reinforcement learning agents request a large number of samples for convergence. Thus, it is necessary to expand datasets to cover more cases. In our experiments, the capacity of both $R_\mathrm{positive}$ and $R_\mathrm{negative}$ pools is set from 20,000 frames to 80,000 frames, sampled by 24 workers.

\section{Experimental Results}

We primarily measure the effectiveness of the proposed ESCE module by adopting different reward functions with DQN and A3C-LSTM agents as the baseline on MountainCar-v0 and six Atari games, respectively. The original A3C-LSTM agent is optimized with environmental rewards only, which is $r_t = 0\cdot r_t^\mathrm{c} + 1\cdot r_t^\mathrm{e}$. Besides, we examine the performance of the agents on new experimental settings, termed as \textit{hindsight rewards} settings, where the environmental rewards is provided with more delayed time. We also examine how purified training process affects the recognition on $R_\mathrm{positive}$ samples and the effectiveness of calibrated rewards with the following two questions:

\begin{itemize}
\item Q1: Does the purified training process help to improve the performance of agents?
\item Q2: How do calibrated rewards affect the convergence of RL training?
\end{itemize}

We attempt to answer Q1 in Sections~\ref{experiment: precision} and~\ref{experiment: identification} and answer Q2 in Sections~\ref{experiment: calibration} and~\ref{experiment: rewards function}.

\begin{figure*}[tp]
\centering
\subfigure[ ] {
 \label{fig:mounntaincar}     
\includegraphics[width=0.8\columnwidth]{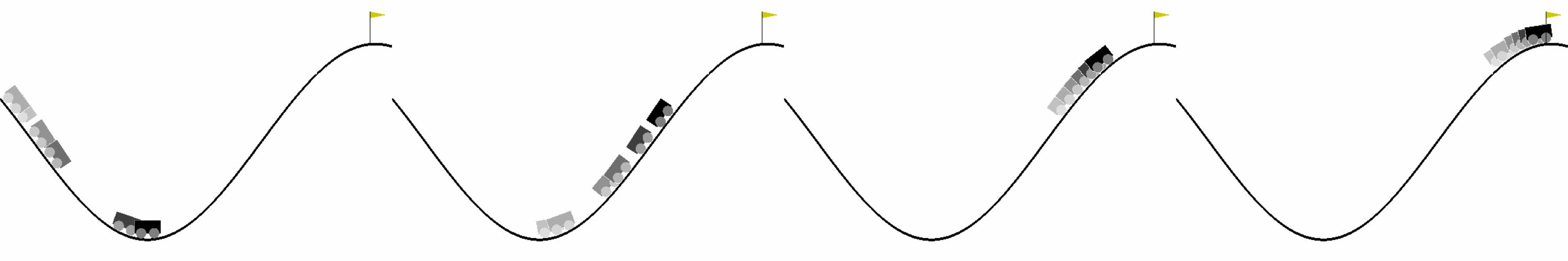} 
}  

\subfigure[ ] { 
\label{fig:mount_results}     
\includegraphics[width=1\columnwidth]{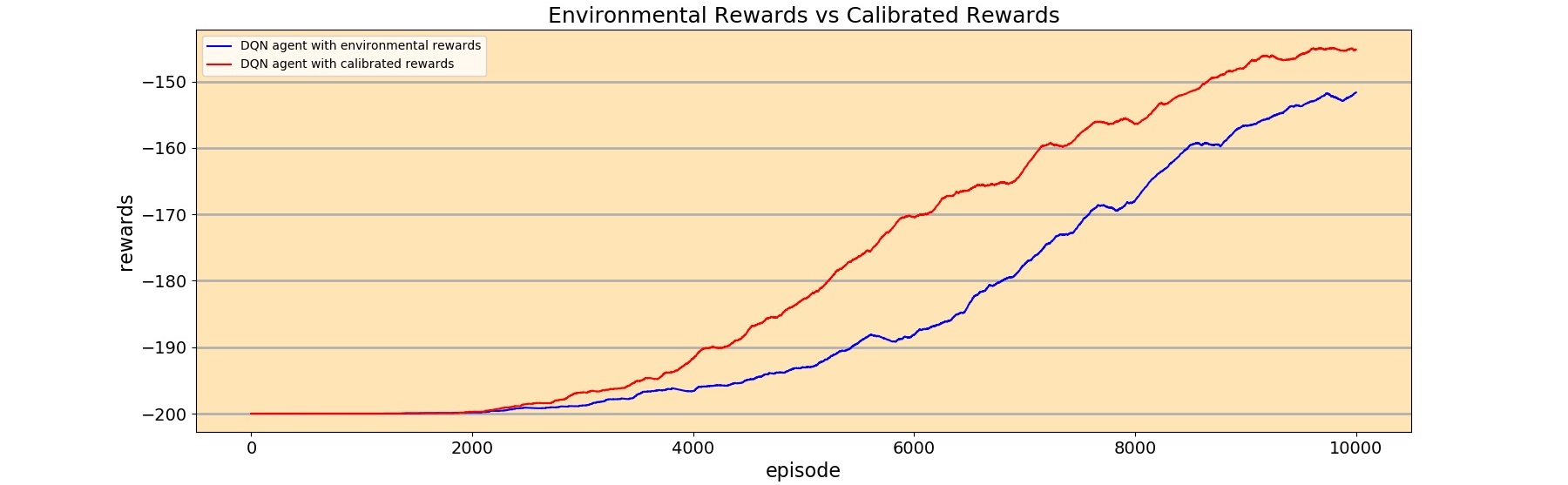}     
}  
\caption{\ref{fig:mounntaincar} exhibits all identified states from one single episode, which record the car drives up the left ramp and accumulate enough potential energy and the rushes to the right at one fling. We stack the frames together to demonstrate the trace, where images of the car fade over time. In \ref{fig:mount_results}, red and blue lines show the average score acquired by DQN agents optimized with calibrated rewards and environmental rewards respectively. Calibrated rewards-driven agents show to learning policy towards target faster.}
\label{fig11}     
\end{figure*}

\subsection{Precision and Recall of Empirical Sufficient State Training}
\label{experiment: precision}

To answer Q1, we first examine the training process of empirical sufficient extraction from a statistical perspective. The {\em Recall} and {\em Precision} of positive samples are a pair of trade-off. Since the $R_\mathrm{positive}$ and $R_\mathrm{negative}$ samples are densely mixed, the purified training process may excludes some $R_\mathrm{positive}$ samples when eliminates negative samples. Empirically, we fine-turned the hyper-parameter $\sigma$ from 0.81 to 1, which in turn keeps both of the {\em Recall} and the {\em Precision} values at a high level. The changes of these two indices are recorded in the right sides of Tables~\ref{table:3} and \ref{table:1}. We also identify that {\em Precision} is positive correlated with the improvement of the policy. This is caused by the random initialization of the policy networks, the erratic performance of the agent makes the reward prediction inaccurate. In most games, both {\em Recall} and {\em Precision} could reach high values after the convergence of the policy networks.

For Breakout-v0 shown in Table~\ref{table:1}, the {\em Recall} is significantly reduced after the agent gets more than 40 marks on average as the bricks hit by the pellet could be vastly different in every episode. Thus, there are larger variances of Breakout-v0 when compared to other games.

\subsection{The Identification of Empirical Efficient States}
\label{experiment: identification}

To further verify the correctness of ESCE, we screenshot the extracted states on three popular games. We empirically find that the identified states have high correlations with rewards acquisition and they are visually similar to the perceptions of humans. In MountainCar-v0, agents need to drive the car uphill to the top, marked by a yellow flag. However the car's engine is not powerful enough to drive the vehicle up to the top in a single pass, therefore the car has to climb up the left ramp to get enough potential energy for later accelerating. We find that ESCE could accurately capture crucial states and provide calibrated rewards to DQN agents only after a few dozens of episodes. Surprisingly, most frames are captured when the car speeds up from a high place on the left ramp to the right. We display stacked frames from one single episode in Figure \ref{fig:mounntaincar}. Furthermore, the results show that the rewards after calibration from ESCE could lead agents to learn target policy faster. We show a comparison of average scores in Figure \ref{fig:mount_results}. 

\begin{figure*}[tp]
\centering
\subfigure[ ] {
 \label{fig:Quantitative Difference}     
\includegraphics[width=1\columnwidth]{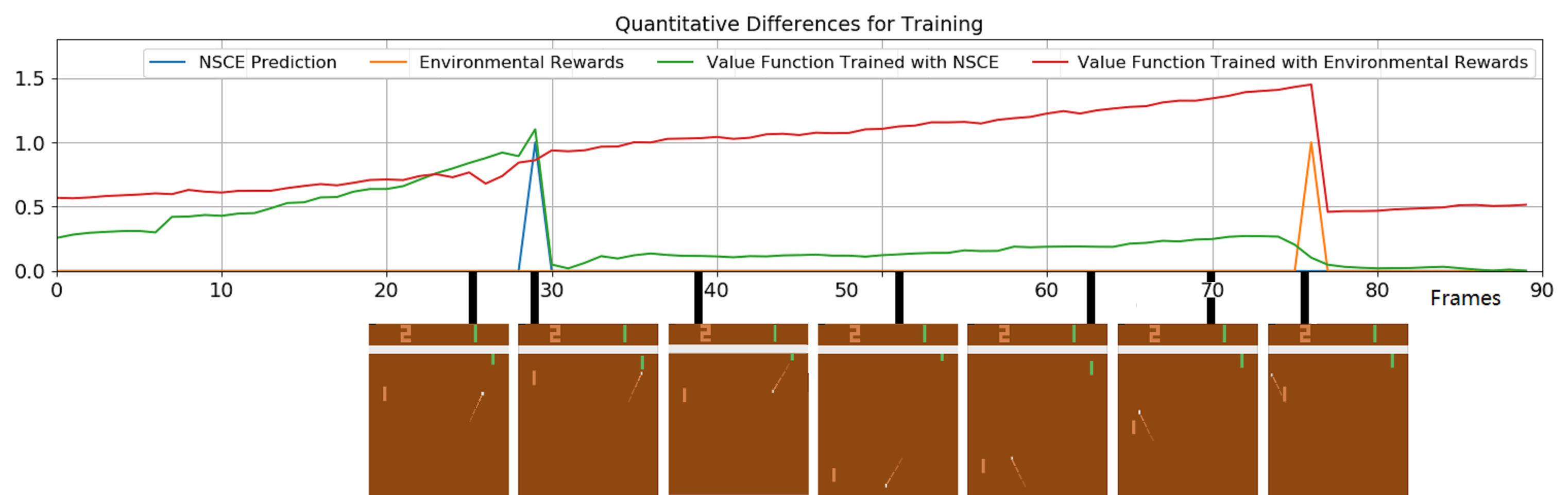}  
}  
\subfigure[ ] {
 \label{fig:d}     
\includegraphics[width=0.31\columnwidth]{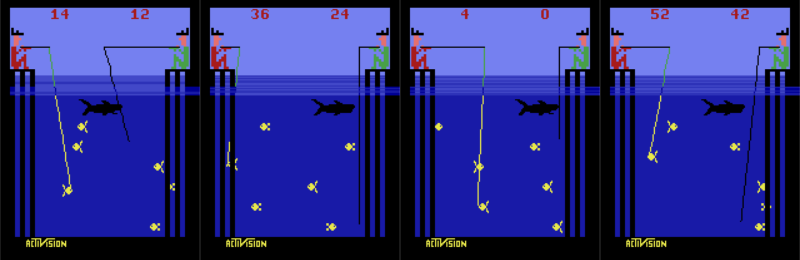}  
}     
\subfigure[ ] { 
\label{fig:e}     
\includegraphics[width=0.31\columnwidth]{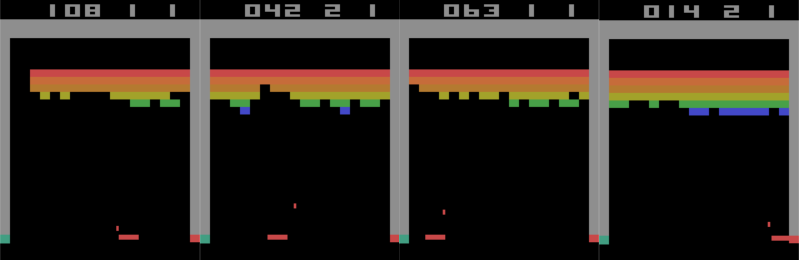}     
}    
\subfigure[ ] { 
\label{fig:f}     
\includegraphics[width=0.31\columnwidth]{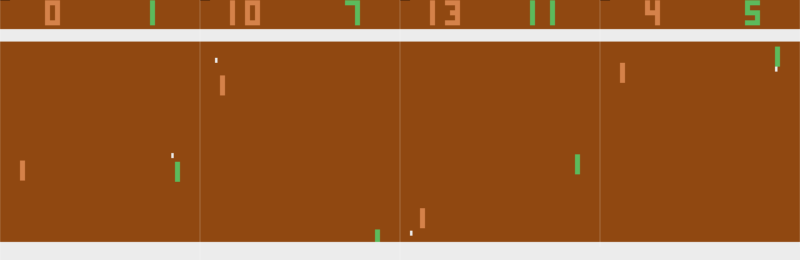}     
}   
\caption{\ref{fig:Quantitative Difference} The calibrated rewards and environmental rewards are presented in blue and yellow lines; the value estimation (Critic) trained with calibrated rewards and environmental rewards are denoted by green and red lines, respectively. The blue line displays that ESCE identifies a empirical sufficient state when the agent hit the pellet with the edge of the bat (right bat). It significantly increases transverse velocity for the ball which lead the agent to win the game. However, the value of baseline (red line) increases continuously until receiving the reward returned by the environment, which is far away from the decisive state. Existing approaches could not make such precise prediction (blue) on critical states. \ref{fig:d} In FishingDerby-v0, most states are identified as the empirical sufficient states when the hook is close to fishes or a fish is already hooked. \ref{fig:e} For Breakout-v0, most identified states occur when the pellet is close to the bat or the bat is on the pellet's potential trajectory. \ref{fig:f} For Pong-v0, the recognized states show that the opponent is about to miss, or agents hit the pellet with the edge of the bat to give the pellet a quick vertical velocity to win the game.}
\label{fig:fig3}     
\end{figure*}

With the evolving of the policy performance, ESD keeps improving as well. In the initial episode, the empirical sufficient checkpoints are recognized a few steps away from the states where the actual rewards are given; with the policy becoming stronger and stabler, more and earlier states can be identified by ESCE. We thus infer that a well-trained ESCE model is capable of making accurate reward predictions through the understanding of the policy and environment with the training goes on. We also visualize the screenshots of extracted states in Figure~\ref{fig:fig3}.

\subsection{Delayed Reward Calibration in Stricter Hindsight Rewards Setting}
\label{experiment: calibration}

The latency of rewards in realistic environments is highly unpredictable. In the reward delay environments, experimental rewards are unable to reflect the performance of policies. Therefore, we further modify the environments to force it to offer even more delayed rewards, termed as hindsight rewards. In this setting, rewards are only provided if an episode ends or after a negative environmental signal. We compare the performance of agents in three games trained with environmental rewards only or combined with the proposed calibrated rewards in the hindsight rewards settings. The results are shown in Table~\ref{table:3}. In this modified scenario, agents guided by environmental rewards only can hardly make any progress, whereas the ones updated with calibrated rewards are able to learn distinctive policies.

The reason behind this phenomenon is that the ESCE model helps to calibrate the delayed reward by identifying those states that meet the empirical sufficient condition, which in turn speeds up the convergence of the policy networks. Given that the state value is gained from the acquired rewards, rewards received from a delayed state would thus mislead the value estimation. Figure~\ref{fig:Quantitative Difference} illustrates the difference of agent value estimation with environmental rewards and calibrated rewards. 

\subsection{Semi-Calibrated and Fully-Calibrated Rewards settings}
\label{experiment: rewards function}

In addition to the hindsight rewards settings, we also conduct experiments on two reward settings to explore how the calibrated rewards affect the training process of a RL agent. The first is the semi-calibrated rewards setting, where the calibrated reward coefficient is set to 0.3 and the environmental reward coefficient is 1, written as $r_t = 0.3\cdot r_t^c + 1\cdot r_t^e$. The second is the fully-calibrated rewards setting, where the reward function is $r_t = 1\cdot r_t^c + 0\cdot r_t^e$. The results are shown in Table~\ref{table:1} as well.

For the semi-calibrated rewards setting, the results show that a small number of calibrated rewards is able to accelerate the training process of agent. It is also can be observed that the higher {\em Recall} and {\em Precision} values are essential to ensure the effectiveness of calibrated rewards. In Breakout-v0 and Boxing-v0, the {\em Recall} can barely reach a high value, and we believe this is due to the large variance in the states of these games (refer to Section~\ref{experiment: identification}).

For the fully-calibrated rewards setting, only calibrated rewards are provided. This ablation study examines the rationality of the time of awarding. The results show that agents trained with calibrated rewards can beat our baseline model in multiple different games. To acquire rewards in FishingDerby-v0, agents need to move a hook to catch fishes. Then, reel back the line before a shark eats the fish. It is common that the shark steals the fish if the agents have not learned to pull the hook up quickly. In the first column of Table~\ref{table:1}, the calibrated rewards significantly boost the convergence speed of the model. Thus, a spike shows. This may be because when the agent learns a stable policy to pull the hook up, the states of a simple action to hook the fish becomes the empirical sufficient state and the calibrated rewards can be provided even earlier. 
This training process also shares similarity of the learning curve of humans, by breaking down complex missions into easy sub-tasks.

\begin{table*}[tp]
    \centering
    \caption{Model convergence comparison of different reward function. The agents in row-one is equipped $r_t = 0.3\cdot r_t^c + 1\cdot r_t^e$ and the agent in row-two is equipped with $r_t = 1\cdot r_t^c + 0\cdot r_t^e$. The real time {\em Precision} and {\em Recall} of $R_\mathrm{positive}$ are presented in the right side of each sub-figure.}
    \label{table:1}
    \includegraphics[scale=0.25]{figures/fig03.png}
    \begin{tabular}{ m{1.91cm}<{\centering} m{1.91cm}<{\centering} m{1.91cm}<{\centering} m{1.91cm}<{\centering} m{1.91cm}<{\centering} m{1.91cm}<{\centering}}
        \toprule
         \tabincell{c}{\scriptsize FishingDerby} & \tabincell{c}{\scriptsize Breakout} & \tabincell{c}{\scriptsize Pong}  & 
         \tabincell{c}{\scriptsize Boxing} & \tabincell{c}{\scriptsize Asterix} & \tabincell{c}{\scriptsize Bowling} \\
        \midrule
        \centerline{\includegraphics[scale=0.092]{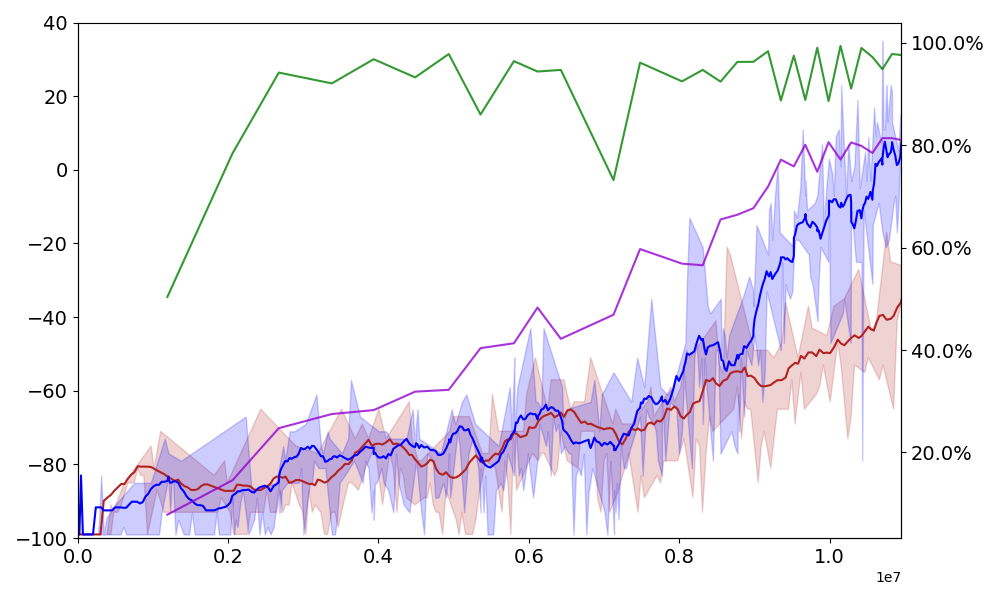}}&
        \centerline{\includegraphics[scale=0.092]{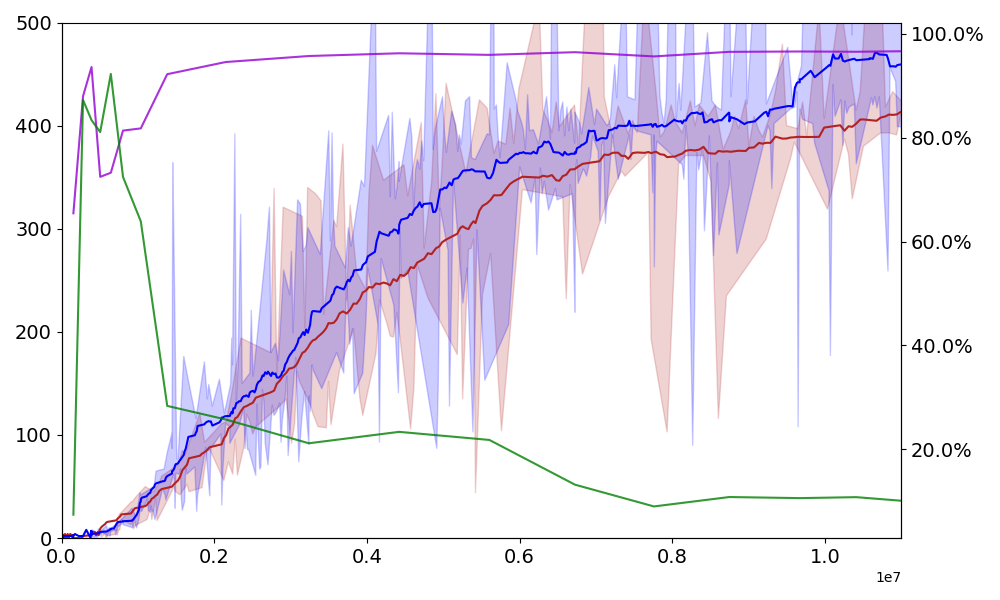}} & 
        \centerline{\includegraphics[scale=0.092]{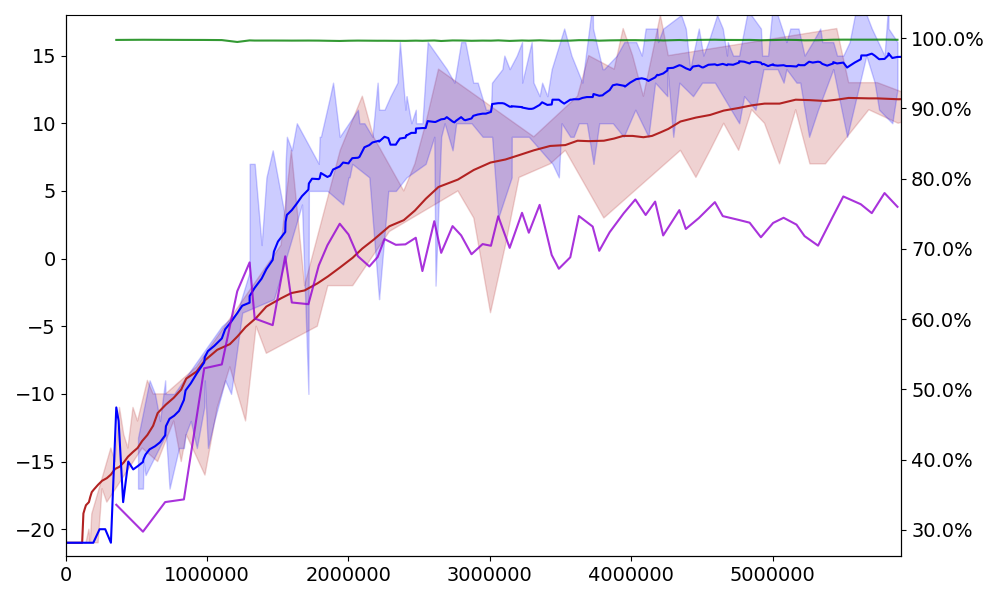}} &
        \centerline{\includegraphics[scale=0.092]{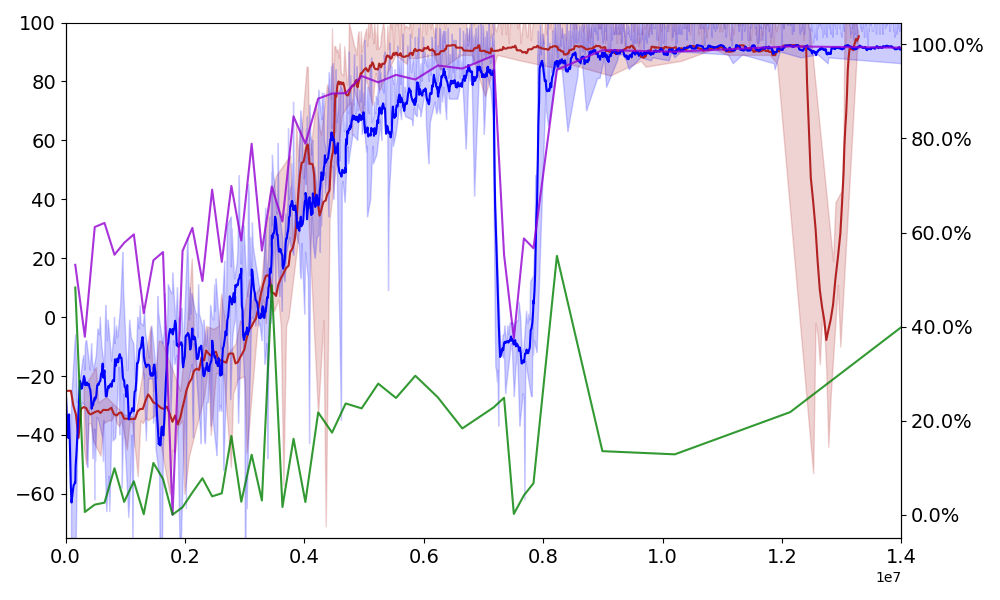}} & \centerline{\includegraphics[scale=0.092]{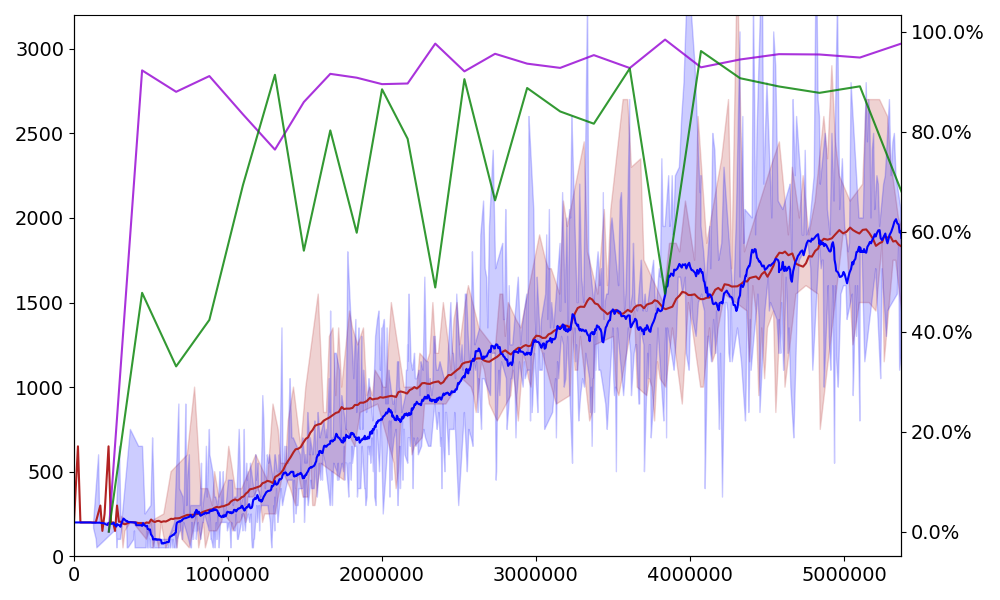}} & \centerline{\includegraphics[scale=0.092]{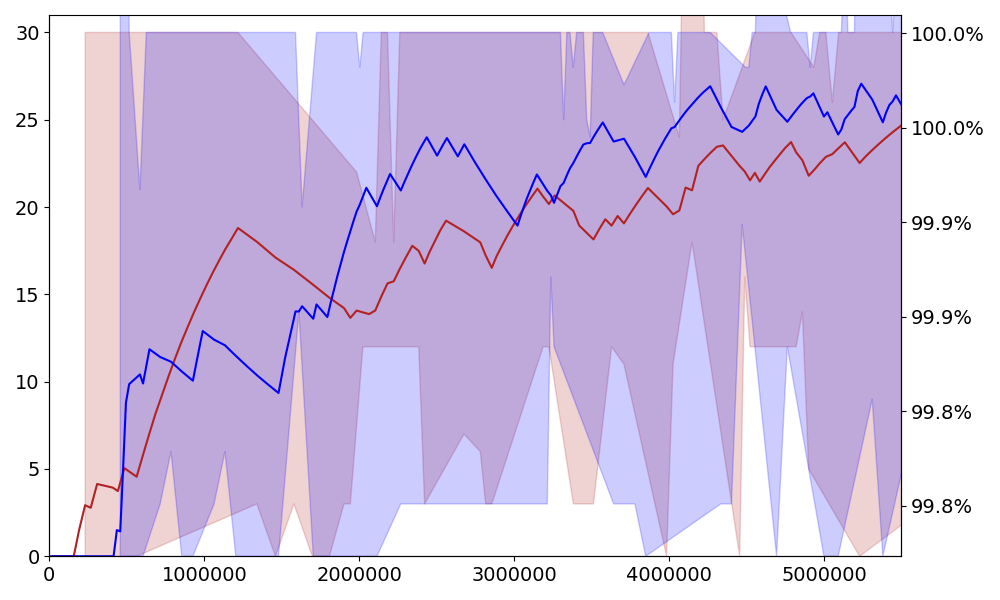}} \\

        \centerline{\includegraphics[scale=0.092]{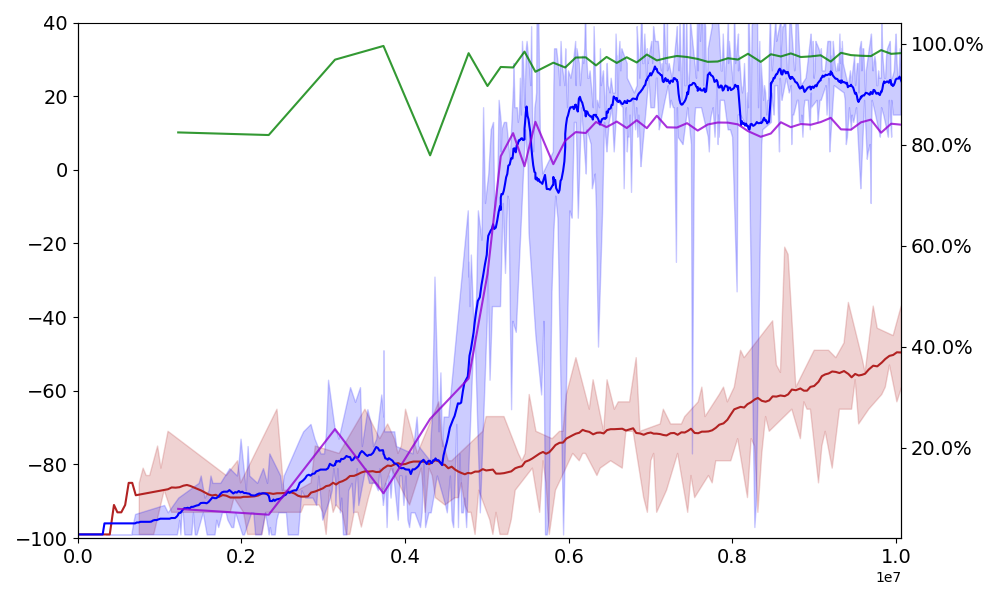}} & 
        \centerline{\includegraphics[scale=0.092]{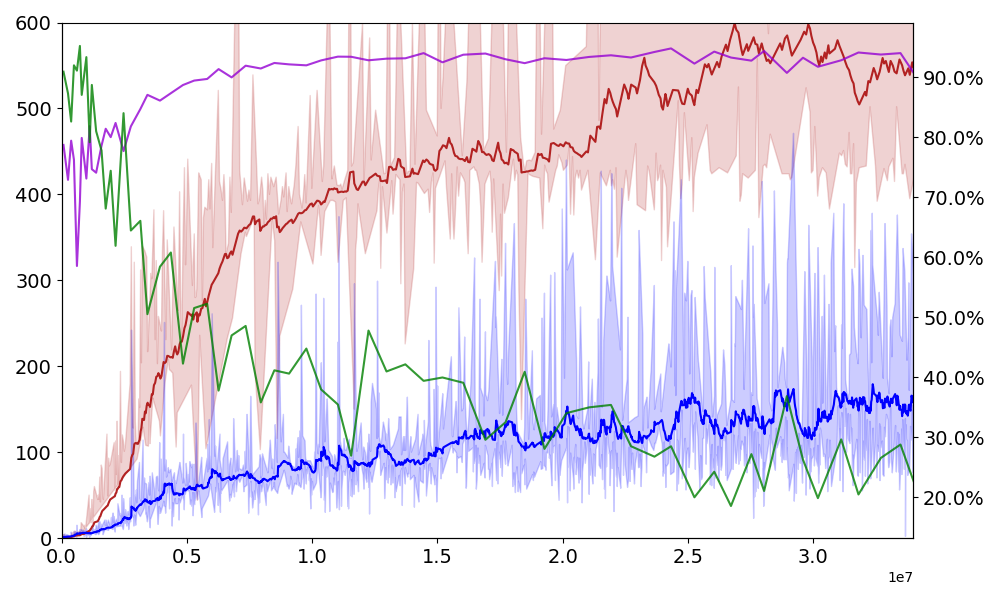}} & 
        \centerline{\includegraphics[scale=0.092]{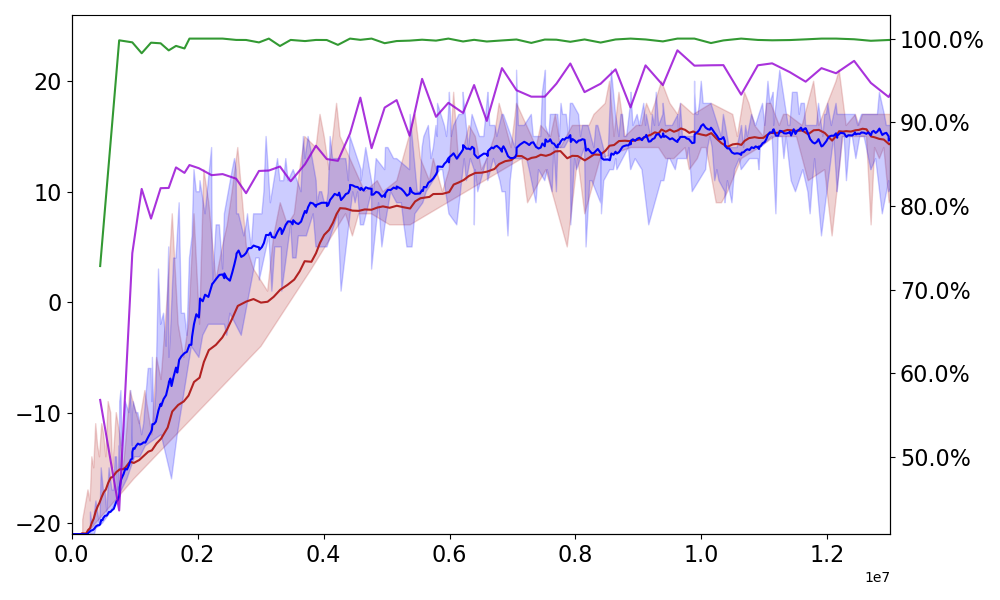}} &
        \centerline{\includegraphics[scale=0.092]{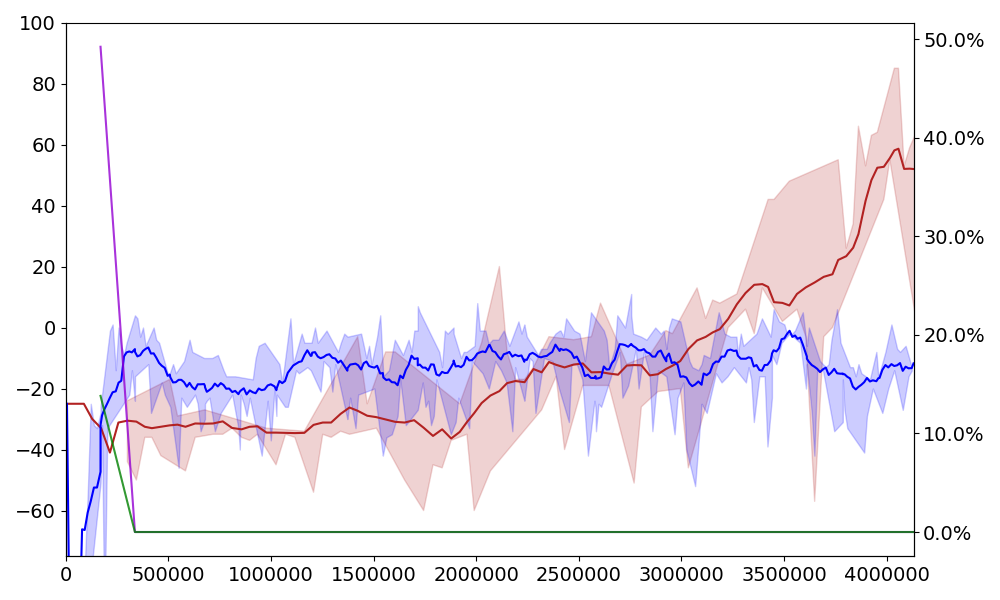}} & \centerline{\includegraphics[scale=0.092]{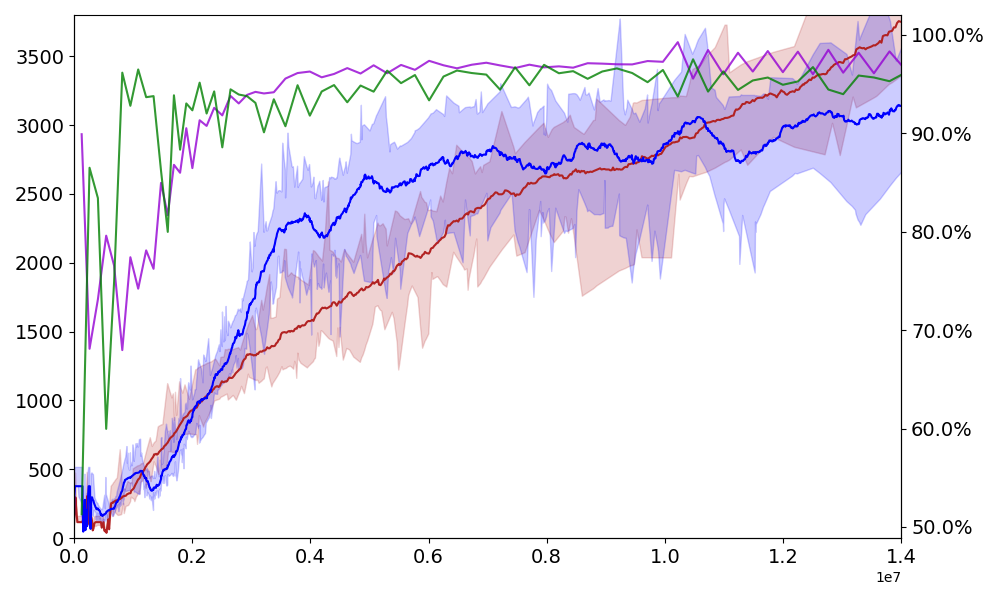}} & \centerline{\includegraphics[scale=0.092]{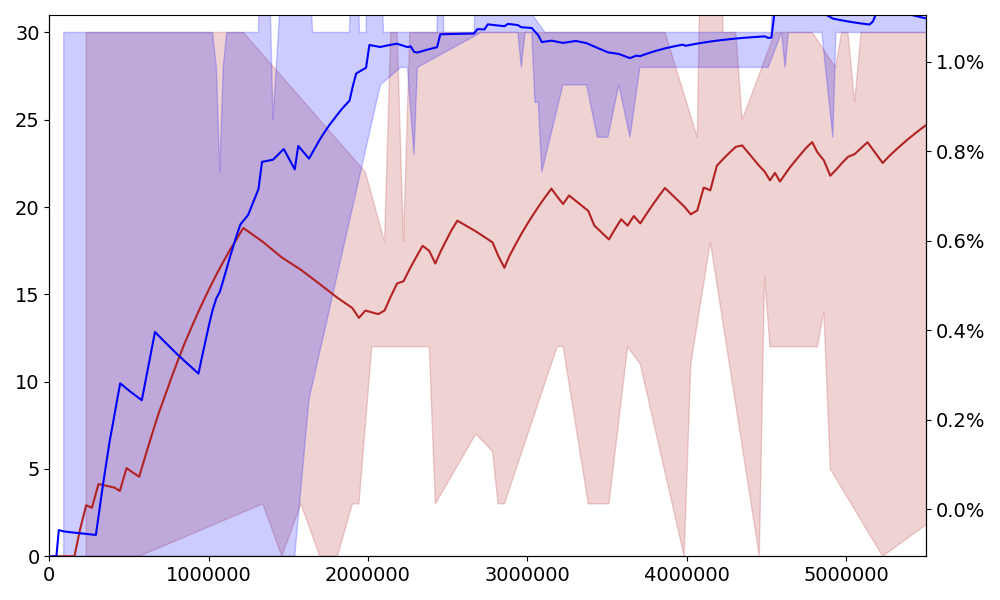}} \\
        \bottomrule
    \end{tabular}
\end{table*}

\section{Conclusion}

In this paper, we formulate an approach to calibrating delayed rewards from a classification perspective. Due to the purified training, the proposed ESCE model is capable of accurately extracting the critical states. Accordingly, the agents trained with calibrated rewards could be assigned with reward without delay. In addition, further experimental results show that agents trained with the proposed calibrated rewards could learn distinctive policies in environments even with extremely delayed rewards. We further identify and discuss the sufficient states extracted by our model resonate with the observations of humans.


\printbibliography
\end{document}